%% file: main_arxiv.tex
\documentclass[11pt,twocolumn]{article}

\usepackage[margin=1in]{geometry}
\usepackage{amsmath,amsfonts,amssymb}
\usepackage{algorithmic}
\usepackage{algorithm}
\usepackage{array}
\usepackage{booktabs}
\usepackage{graphicx}
\usepackage{multirow}
\usepackage{float}
\usepackage{colortbl}
\usepackage{xcolor}
\usepackage{url}
\usepackage[numbers,sort&compress]{natbib}
\usepackage{hyperref}
\usepackage[capitalize]{cleveref}
\usepackage{docmute}

\hypersetup{
    colorlinks=true,
    linkcolor=blue!70!black,
    citecolor=blue!70!black,
    urlcolor=blue!70!black
}

\title{FrontierGS: Progressive View-Space Frontier Expansion for Sparse 3D Gaussian Splatting}
\author{
    Zijian Wu$^{1}$, Mingfeng Jiang$^{1,\ast}$, Zidian Lin$^{1}$, Ying Song$^{1}$,\\
    Ziqian Lu$^{1}$, Qun Wu$^{2}$, and Hanjie Ma$^{1}$\\[0.5em]
    \small $^{1}$School of Artificial Intelligence, Zhejiang Sci-Tech University, Hangzhou 310018, China\\
    \small $^{2}$School of Art$\cdot$Design, Zhejiang Sci-Tech University, Hangzhou 311103, China\\[0.25em]
    \small $^{\ast}$Corresponding author: Mingfeng Jiang (m.jiang@zstu.edu.cn)
}
\date{}

\begin{document}
\maketitle

\bibliographystyle{plainnat}
\makeatletter
\newenvironment{frontmatter}{}{}
\newenvironment{keyword}{\begin{center}\small\textbf{Keywords:} }{\end{center}}
\newcommand{\sep}{\quad}
\def\title#1{}
\def\author#1{}
\providecommand{\journal}[1]{}
\renewcommand{\bibliographystyle}[1]{}
\makeatother

\input{main_anonymous}

\end{document}

%% file: main_anonymous.tex
\begin{frontmatter}

\title{FrontierGS: Progressive View-Space Frontier Expansion for Sparse 3D Gaussian Splatting}

\author{Anonymous Submission}

\begin{abstract}
3D Gaussian Splatting (3DGS) enables efficient, high-fidelity novel view synthesis, yet its performance degrades severely under sparse-view supervision due to extreme ill-posedness. While existing approaches introduce various geometric priors or heuristic regularizations to mitigate overfitting, they largely overlook the underlying optimization dynamics. In this paper, we identify a distinct training pathology termed view-space generalization collapse: as optimization proceeds, observed-view fitting error continuously decreases, whereas rendering quality at unseen viewpoints initially peaks at an intermediate stage and subsequently deteriorates. This transient peak indicates that early-stage optimization implicitly uncovers valid local 3D structures before overfitting along training rays destroys them. Grounded in this finding, we propose FrontierGS, a progressive view-space expansion framework designed to capture and anchor these transiently reliable states. Operating along an expanding generalization frontier, FrontierGS systematically probes candidate viewpoints around observed cameras, verifies their geometric and perceptual consistency, and progressively incorporates reliable candidates into the supervision set to permanently preserve valid geometry before collapse occurs. Extensive experiments across the LLFF, MipNeRF-360, and DTU benchmarks demonstrate that FrontierGS effectively eliminates view-space collapse, consistently outperforming state-of-the-art baselines in both rendering fidelity and geometric accuracy.
\end{abstract}


\end{frontmatter}

\section{Introduction}
Rapid advances in 3D scene reconstruction and novel view synthesis have established foundational technologies for immersive virtual reality, digital twin modeling, cultural heritage preservation, robotics, and industrial inspection \cite{yu2024sensors, tao2022digital, skublewska20223d, mgss, LIANG2026133170}. For years, Neural Radiance Fields (NeRF) \cite{mildenhall2021nerf} and their continuous volumetric extensions \cite{barron2021mip, somraj2023vip, chen2021mvsnerf, pumarola2021d} set the benchmark for photorealistic visual fidelity. However, the implicit continuous formulations of NeRF suffer from sluggish optimization trajectories and computationally demanding ray-marching inference. Recently, 3D Gaussian Splatting (3DGS) \cite{kerbl20233d} revolutionized the field by combining explicit, point-based anisotropic Gaussian primitives with a tile-based, differentiable rasterization pipeline. This representation delivers high-fidelity rendering with real-time performance, sparking widespread adoption across diverse computer vision and graphics domains \cite{qian20243dgs, liu2024citygaussian, wang2024view, mallick2024taming, zhang2025wheat3dgs, szymanowicz2024splatter, clocap, plgs}.

Despite its remarkable efficacy under dense multi-view captures, 3DGS suffers a precipitous drop in performance when input viewpoints are highly sparse \cite{long2022sparseneus, han2024binocular, cai2024structure, xu2024grm, charatan2024pixelsplat}. Sparse-view reconstruction poses a severe challenge due to its highly ill-posed nature. Wide baselines and limited visual overlap create substantial supervision voids across the camera pose manifold, leading standard 3DGS optimization to produce stretched Gaussian primitives, persistent floating artifacts, and severe structural collapse when rendered from unobserved trajectories.

To counter data scarcity in sparse-view reconstruction, prior research has largely pursued two main avenues. The first direction incorporates external geometric priors, such as monocular depth or normal cues, alongside hand-crafted regularization terms to constrain primitive optimization \cite{wang2023sparsenerf, zhu2024fsgs, li2024dngaussian, bao2025loopsparsegs}. Although these auxiliary constraints help suppress floating artifacts, they operate primarily through passive, static boundaries rather than addressing the underlying optimization dynamics of sparse 3DGS. As a result, such methods offer limited generalization capacity when rendering novel viewpoints, while also remaining vulnerable to scale inconsistencies and domain shifts in monocular estimations. The second direction leverages pretrained 2D or multi-view diffusion models to synthesize pseudo-views for data augmentation \cite{11095131, 10657150}. However, generative diffusion models frequently introduce unconstrained hallucinations that violate 3D structural consistency, causing visual blurring, geometric drift, and substantial training overhead.

\begin{figure*}[!t] 
    \centering    
    \includegraphics[width=1\linewidth]{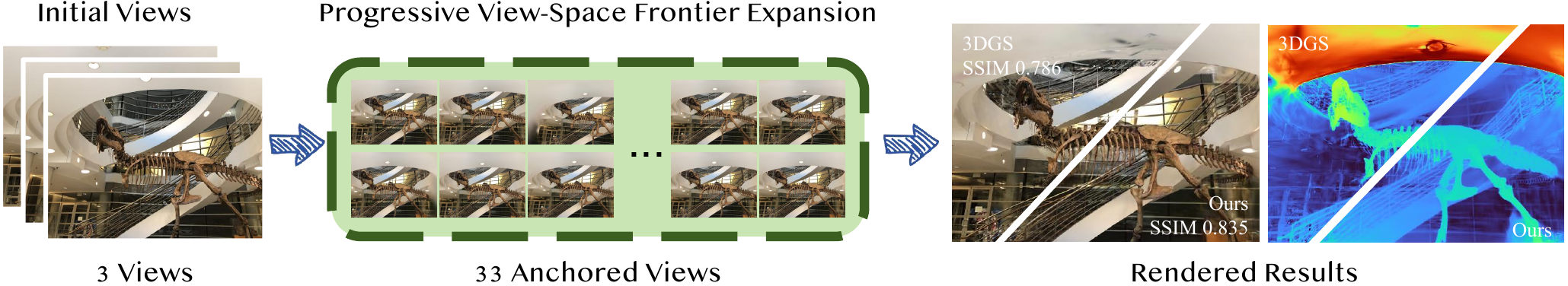} 
    \caption{Novel-view synthesis comparison under sparse-view supervision. Standard 3DGS suffers from view-space generalization collapse, leading to severe visual artifacts. In contrast, FrontierGS progressively probes and anchors reliable novel viewpoints, yielding photorealistic and artifact-free renderings.}
    \label{fig:intro}
\end{figure*}

\begin{figure}[!t]
    \centering    
    \includegraphics[width=1\linewidth]{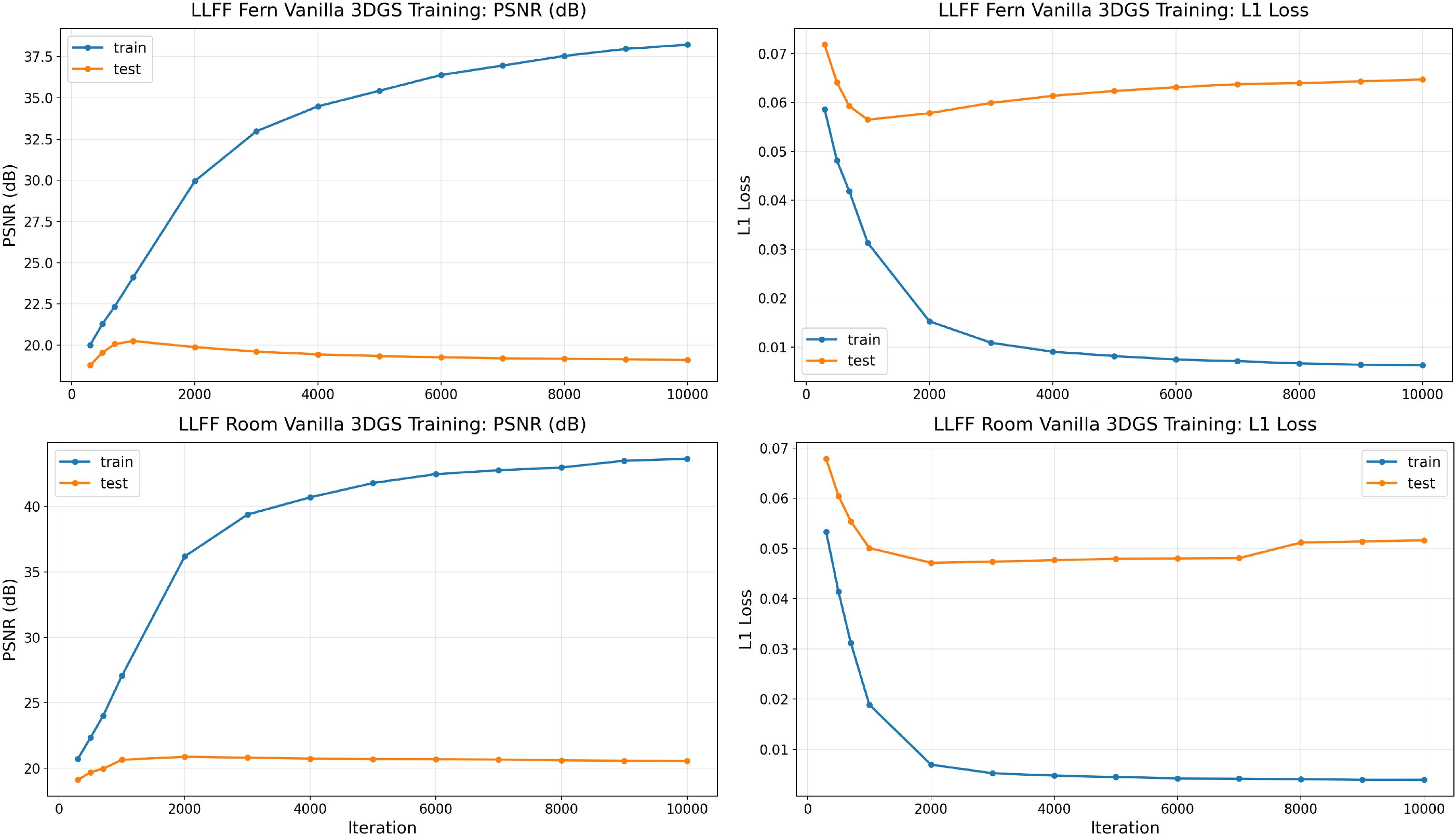}
    \caption{Training curves of vanilla 3DGS under sparse-view supervision on the \textit{Fern} and \textit{Room} scenes from the LLFF dataset. While training loss on observed views continuously decreases, novel-view rendering quality initially peaks at an intermediate stage and subsequently degrades, empirically demonstrating the view-space generalization collapse pathology.}
    \label{fig:collapse_dynamics}
\end{figure}

In this work, we conduct an in-depth investigation into the intrinsic optimization process and underlying failure modes of sparse-view 3D reconstruction. Conventionally, the fundamental dilemma in sparse-view reconstruction is broadly summarized as overfitting to the scarce observed viewpoints, which inherently hampers novel-view generalization \cite{LIU2026129334, zhu2024fsgs, li2024dngaussian}. However, what this overfitting physically entails in explicit 3D representations, where it actually originates, and how it dynamically evolves throughout training have rarely been deeply analyzed. Through systematic empirical tracking, we discover an intriguing optimization phenomenon that extends beyond conventional understandings of overfitting. While photometric fitting on the sparse observed viewpoints monotonically improves throughout training, the reconstruction quality at unseen viewpoints does not follow a corresponding upward trend. Instead, novel-view rendering fidelity consistently reaches a distinct peak at an intermediate training stage and subsequently deteriorates as optimization continues, a behavior illustrated in Fig.~\ref{fig:collapse_dynamics}. This observation reveals a critical insight that during the training process, the model actually reaches a locally optimal or relatively reliable state at unobserved viewpoints. However, owing to the absence of ground-truth supervision, this transient state is neither continuously reinforced nor preserved, but instead gets progressively disrupted as optimization continues. This degradation trajectory precisely embodies the true manifestation of overfitting in sparse-view reconstruction, which we formalize as view-space generalization collapse.

Grounded in this insight, we propose FrontierGS, a progressive view-space expansion framework that shifts the paradigm from passive static constraints or unconstrained generative synthesis to actively capturing and anchoring transiently reliable optimization states. The core intuition is grounded in our dynamic analysis: since sparse 3DGS temporarily constructs valid geometry at unobserved viewpoints before line-of-sight overfitting degrades it, identifying and anchoring these reliable states along the camera manifold can establish trusted intermediate steps for outward expansion. Specifically, FrontierGS maintains a dynamically advancing generalization frontier that systematically expands from observed cameras toward unobserved regions. Along this expanding frontier, the framework periodically probes candidate viewpoints using an adaptive exploration radius. To stabilize intermediate optimization, these probed views are structurally regularized through Pearson depth correlation and dual-model consensus constraints. Concurrently, candidate viewpoints are tracked using a multi-signal quality assessment that integrates structural similarity, perceptual realism, and no-reference visual metrics. Once verified as structurally sound and visually coherent, these views are permanently incorporated into the active supervision set to safeguard valid 3D geometry against subsequent overfitting. As local regions solidify, the exploration radius progressively widens, pushing the generalization frontier further into unobserved camera space.

The main contributions of this work are summarized as follows:
\begin{itemize}
    \item We reveal the pathology of view-space generalization collapse in sparse 3DGS, identifying that novel-view representations transiently reach reliable states before overfitting causes structural degradation.
    \item We propose FrontierGS, a progressive view-space expansion framework that actively probes, verifies, and anchors transiently reliable states along a dynamically advancing generalization frontier.
    \item We design a joint regularization and verification mechanism, utilizing depth-guided consensus to stabilize optimization and a multi-signal quality metric to filter high-confidence anchor views.
   \item Extensive experiments across standard benchmarks demonstrate that FrontierGS outperforms state-of-the-art methods in both rendering fidelity and geometric reconstruction accuracy.
\end{itemize}

\begin{figure*}[!t]
    \centering    
    \includegraphics[width=1\linewidth]{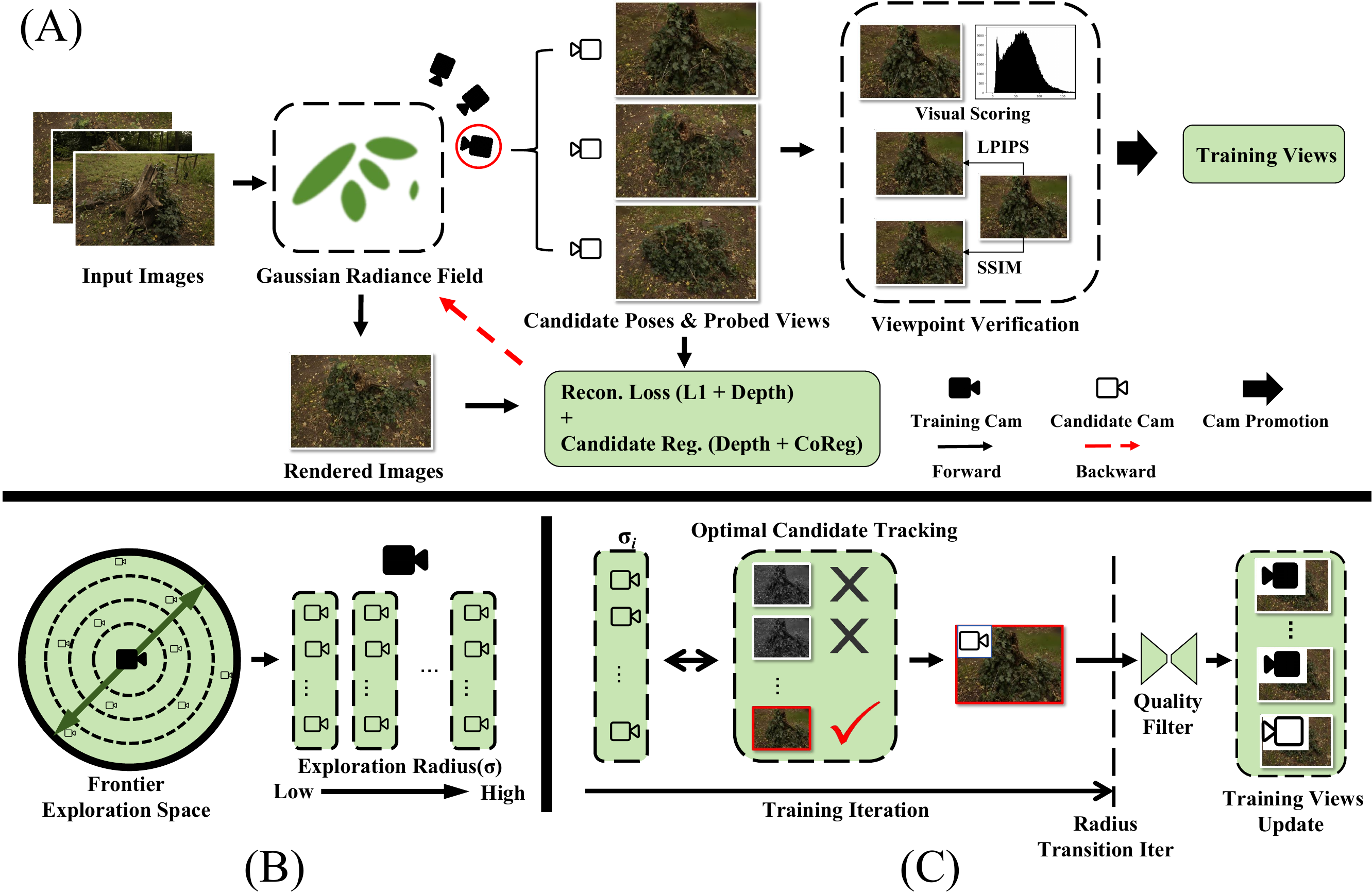}  
    \caption{The overall architecture of the FrontierGS framework is shown in (A). The pipeline consists of three key stages: (1) candidate view probing, where virtual camera poses are sampled around observed viewpoints across expanding exploration radii, as detailed in (B); (2) progressive frontier scheduling, which gradually advances the exploration radius during training as local geometry stabilizes; and (3) candidate view evaluation and anchoring, where each candidate is scored using perceptual (LPIPS), structural (SSIM), and no-reference quality metrics while being regularized by depth correlation and dual-model consensus. Only the most reliable candidate at each radius level that satisfies the quality criterion is permanently anchored into the training set, as illustrated in (C). This progressive process preserves sound intermediate 3D representations and prevents view-space generalization collapse.}
    \label{fig:pipeline}
\end{figure*}

\section{Related Work}

\subsection{Neural Radiance Fields and 3D Gaussian Splatting}
NeRF \cite{mildenhall2021nerf} parameterized continuous 3D scenes as multilayer perceptrons (MLPs) mapping 5D spatial coordinates and viewing directions to volume density and color. Subsequent works improved NeRF's anti-aliasing \cite{barron2021mip}, dynamic modeling \cite{pumarola2021d}, and generalizability \cite{chen2021mvsnerf}. However, the continuous volumetric rendering equation requires evaluating hundreds of MLP queries along each ray, resulting in prolonged optimization times and rendering frame rates unsuitable for interactive applications.

To overcome these computational bottlenecks, 3DGS \cite{kerbl20233d} introduced an explicit radiance field representation utilizing millions of parameterized 3D Gaussians. Each primitive possesses a 3D mean position, full covariance matrix parameterized by scaling and rotation quaternions, opacity, and spherical harmonic (SH) coefficients. By employing an efficient tile-based, depth-sorted differentiable rasterizer, 3DGS achieves real-time rendering at 100+ FPS while retaining photorealistic fidelity. Recent extensions have scaled 3DGS to city-scale environments \cite{liu2024citygaussian, FEI2026130758}, dynamic capture \cite{szymanowicz2024splatter}, and compressed representations \cite{mallick2024taming}. Nevertheless, because 3DGS relies on explicit spatial primitives supervised strictly by 2D screen-space photometric gradients, its optimization becomes acutely vulnerable when supervisory viewpoints are scarce.

\subsection{Sparse-View 3D Reconstruction}
Reconstructing 3D scenes from sparse observations (e.g., 2--5 viewpoints) is a long-standing challenge in computer vision \cite{schonberger2016structure, wang2018pixel2mesh, qi2017pointnet, long2022sparseneus}. Under sparse inputs, the visual hull and radiance field are severely underconstrained, triggering catastrophic geometric collapse and phantom density artifacts.

\paragraph{NeRF-based sparse regularization}
To stabilize sparse NeRFs, several regularizers have been introduced. RegNeRF \cite{niemeyer2022regnerf} regularizes patches rendered from unobserved viewpoints using appearance and geometric smoothness priors. FreeNeRF \cite{yang2023freenerf} imposes frequency-band scheduling on positional encodings alongside near-camera density penalties, mitigating floater formation without external priors. SparseNeRF \cite{wang2023sparsenerf} distills coarse depth-ranking constraints from pretrained monocular depth estimators to guide density placement. Meanwhile, pixelNeRF \cite{yu2021pixelnerf} conditions radiance fields on multi-view image features across scenes. Although these methods alleviate visual degradation, their implicit volumetric representations remain computationally heavy and frequently produce over-smoothed surfaces in sparse regimes.

\paragraph{3DGS-based sparse regularization}
Adapting 3DGS to sparse settings has attracted intense interest \cite{li2024dngaussian, zhu2024fsgs, zhang2024cor, zheng2025nexusgs, park2025dropgaussian, paliwal2024coherentgs, liu2026sv, zhang2026usgs}. FSGS \cite{zhu2024fsgs} initializes Gaussians from sparse Structure-from-Motion (SfM) points and utilizes monocular depth priors to steer Gaussian growth. DNGaussian \cite{li2024dngaussian} introduces depth-normalization and multi-scale depth loss to regularize Gaussian geometry. DropGaussian \cite{park2025dropgaussian} applies stochastic dropout to Gaussian primitives during training to prevent co-adaptation. CoR-GS \cite{zhang2024cor} trains dual Gaussian models and prunes inconsistent splats via mutual photometric and geometric consensus. NexusGS \cite{zheng2025nexusgs} exploits cross-view epipolar geometry and optical flow constraints. Furthermore, recent methods incorporate semantic priors \cite{tang2025spars3r}, confidence-weighted enhancement \cite{xiong2025sparsegs}, or loop-based densification \cite{bao2025loopsparsegs}.

However, these methodologies primarily operate through static supervision or localized regularizations. They generally overlook the dynamic evolution of camera-space optimization and remain vulnerable to domain shifts and scale inconsistencies from external priors. Consequently, such static or external constraints offer limited control over the dynamic view-space overfitting that progressively intensifies during optimization.

\section{Method}

In this section, we present FrontierGS, a progressive view-space expansion framework for sparse-view 3DGS that systematically identifies, regularizes, and anchors transiently reliable 3D states into the supervision domain. The overall pipeline of the proposed method is illustrated in Fig.~\ref{fig:pipeline}.

\subsection{Preliminaries: Gaussian Splatting}

3DGS represents a scene as a set of anisotropic Gaussian primitives, each parameterized by position $\mathbf{\mu} \in \mathbb{R}^3$, covariance $\Sigma \in \mathbb{R}^{3 \times 3}$, opacity $\alpha \in [0,1]$, and spherical harmonic (SH) coefficients for view-dependent color. Formally, each Gaussian defines a density function in 3D space:
\begin{equation}
G(\mathbf{x}) = \exp\left(-\tfrac{1}{2}(\mathbf{x}-\mathbf{\mu})^\top \Sigma^{-1} (\mathbf{x}-\mathbf{\mu}) \right)
\end{equation}

The rendering process follows a rasterization pipeline, in which each Gaussian is splatted into screen space, composited according to depth order, and accumulated using alpha blending. Specifically, the pixel color $\mathbf{C}(\mathbf{p})$ at screen location $\mathbf{p}$ is obtained by front-to-back alpha compositing:
\begin{equation}
\mathbf{C}(\mathbf{p}) = \sum_{i=1}^N T_i \, \alpha_i \, \mathbf{c}_i(\mathbf{p}),
\quad
T_i = \prod_{j=1}^{i-1} (1-\alpha_j)
\end{equation}
where $\alpha_i$ denotes the opacity of the $i$-th Gaussian, $\mathbf{c}_i(\mathbf{p})$ is its color contribution, and $T_i$ is the transmittance term accounting for accumulated transparency from closer Gaussians.

\subsection{Progressive View-Space Probing and Scheduling}
\begin{figure*}[t]
    \centering    
    \includegraphics[width=1\linewidth]{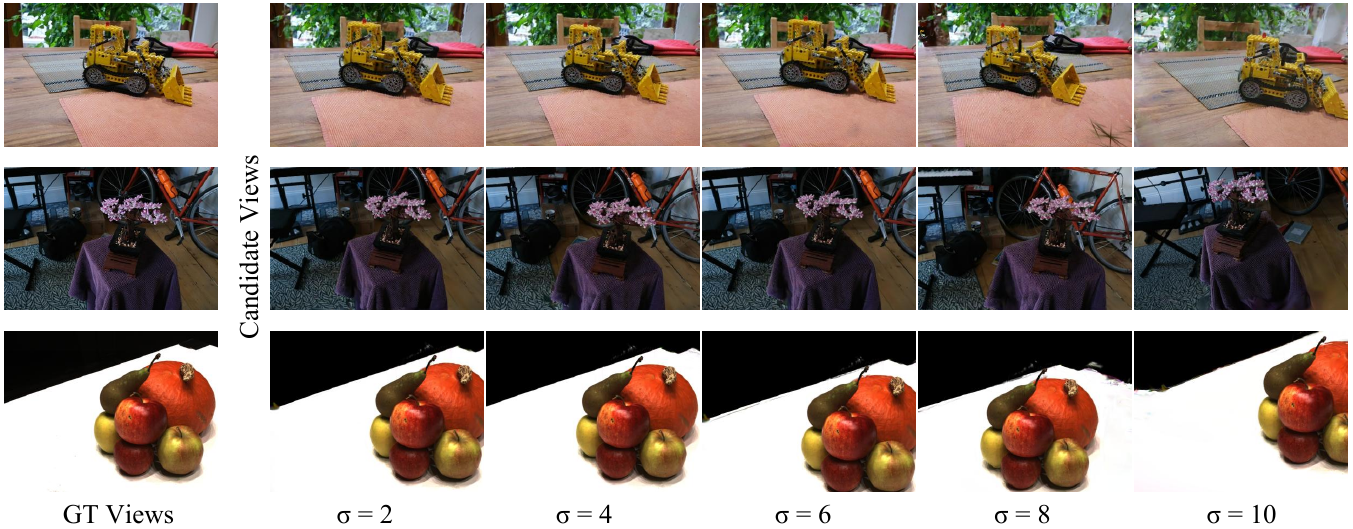}
    \caption{Visualization of candidate view probing. Examples of virtual camera poses probed with different exploration radii $\sigma$ around an observed camera pose.}
    \label{fig:curi}
\end{figure*}

\paragraph{Candidate view probing}
As established in our observation, sparse 3DGS optimization transiently achieves locally optimal and reliable representations in view-space regions adjacent to observed viewpoints before overfitting ruins them. To identify and capture these reliable intermediate states, we systematically probe the local view space around observed cameras. Specifically, candidate viewpoints are generated by perturbing the camera extrinsics of observed views within a controlled exploration radius. Given an observed camera with rotation matrix $R$ and translation vector $T$, its optical center in world coordinates is computed as $C = -R^\top T$. We then apply continuous angular perturbations in yaw and pitch sampled from a zero-mean Gaussian distribution with a standard deviation $\sigma$ (in degrees), coupled with a mild radial translation perturbation $\sigma_r$ along the viewing direction:
\begin{equation}
\begin{aligned}
R' &= R\,R_{\Delta}(\sigma), \quad
C' = C(1 + \epsilon_r), \quad
\epsilon_r \!\sim\! \mathcal{N}(0, \sigma_r^2),\\[4pt]
&\hspace{3.3cm} T' = -R'C'
\end{aligned}
\end{equation}
The perturbed parameters $(R', T')$ define a candidate virtual viewpoint centered in the local neighborhood of the observed camera, preserving the original intrinsic calibration.

For each observed camera $\mathcal{C}_t$, we initialize hierarchical candidate groups $\mathcal{P}(\sigma_i) = \{\mathcal{C}_s^{\sigma_i, j}\}_{j=1}^{N_i}$ across discrete exploration radii $\sigma_i$. Rather than scattering unconstrained virtual views, this structured sampling establishes an orderly progression from tight local neighborhoods where geometry is most stable toward wider unobserved trajectories. Fig.~\ref{fig:curi} illustrates candidate camera poses probed across different exploration radii.

\paragraph{Progressive frontier scheduling}
Instead of probing the entire camera-pose space indiscriminately from the beginning—which would expose the model to severely uncertain regions where 3D geometry has not yet coalesced—we adopt a progressive scheduling strategy that gradually advances the exploration frontier outward. Training commences by probing candidate viewpoints at a minimal exploration radius $\sigma_{\min}$, allowing the model to first consolidate and verify local geometric consistency near the observed viewpoints. Once the inner view space is verified and anchored, the exploration radius is systematically expanded to probe wider viewpoint displacements, progressively moving the generalization frontier into previously unobserved view space.

Formally, the active exploration radius at iteration $t$ is scheduled as:
\begin{equation}
\sigma_{\text{active}}(t) = \min\big(\sigma_{\max}, \, \sigma_{\min} + k \cdot \lfloor t / T_s \rfloor \big)
\end{equation}
where $\sigma_{\min}$ and $\sigma_{\max}$ denote the minimum and maximum exploration radii, respectively. $T_s$ denotes the iteration interval after which the frontier advances to the next radius level, and $k$ is the incremental step between successive stages. By governing the expansion trajectory via $\sigma_{\text{active}}(t)$, the model steadily anchors reliable representations from immediate camera neighborhoods before attempting wider view-space exploration, effectively preventing view-space generalization collapse.

\subsection{Candidate View Evaluation and Anchoring}

To ensure that only genuinely sound and artifact-free representations are admitted into the permanent supervision pool, we formulate a two-stage evaluation and anchoring mechanism. This protocol rigorously monitors candidate fidelity during optimization and selectively anchors the most reliable candidates before overfitting degrades the underlying 3D structure.

\paragraph{Evaluation during training}  
At each optimization iteration, for each observed camera $\mathcal{C}_t$, a candidate viewpoint $\mathcal{C}_s^{\sigma_{\text{active}}, j} \in \mathcal{P}(\sigma_{\text{active}})$ is sampled from the pool corresponding to the current active exploration radius $\sigma_{\text{active}}$. To evaluate the rendering fidelity of the candidate viewpoint in the absence of unobserved ground-truth images, the rendered candidate image is assessed against the reference image of the adjacent observed view using a composite multi-signal metric. This metric jointly incorporates Structural Similarity (SSIM) to evaluate structural coherence, Learned Perceptual Image Patch Similarity (LPIPS) to measure feature-level realism, and a no-reference image-quality evaluator \cite{agnolucci2024qualityaware} to penalize synthetic artifacts, noise, and floating density debris. Throughout the iterations within the current radius stage, we continuously track the candidate that achieves the optimal composite score for each $(\mathcal{C}_t, \sigma_i)$ pair. 

Crucially, during training, candidate viewpoints are simultaneously regularized via auxiliary geometric constraints, namely Pearson depth correlation and dual-model consensus (detailed in \Cref{sec:loss}). These regularizers act directly on the backpropagation gradients to enforce structural consistency, while the multi-signal metric serves as an objective tracking and screening criterion.

\paragraph{Anchoring to training views}  
When the exploration schedule transitions from $\sigma_{\text{prev}}$ to $\sigma_{\text{next}}$, the optimal candidate viewpoint tracked during the previous stage is evaluated for permanent anchoring. To prevent perspective parallax from biasing selection toward poses closest to observed views, we decouple reference comparisons and evaluate the candidate via the no-reference quality score $\mathcal{Q}_{\text{NR}}(\hat{\mathbf{I}}_s^*)$. The quality threshold $\tau$ is adaptively calibrated from the average quality of the $M$ observed ground-truth views:
\begin{equation}
\tau = \frac{1}{M} \sum_{k=1}^M \mathcal{Q}_{\text{NR}}(\mathbf{I}_k^{\text{GT}}) + \delta
\end{equation}
where $\delta$ is a small tolerance margin. Candidates satisfying $\mathcal{Q}_{\text{NR}}(\hat{\mathbf{I}}_s^*) \le \tau$ are permanently anchored into $\mathcal{V}_{\text{train}}$, freezing reliable 3D representations before view-space collapse occurs and providing supervisory support for subsequent outward expansion.

\subsection{Optimization Objective}
\label{sec:loss}
To drive effective scene optimization across the expanding supervision set while strictly preserving fidelity to the physical observations, the overall objective function is formulated with three complementary components:
(1) a dynamic reconstruction loss $\mathcal{L}_{\text{train}}$ defined over the expanding training set (comprising both initial observed views and newly anchored viewpoints);
(2) an anchor loss $\mathcal{L}_{\text{anchor}}$ strictly enforced on the original observed views to eliminate semantic drift; and
(3) geometric regularization terms $\mathcal{L}_{\text{reg}}$ imposed on probed candidate viewpoints, consisting of Pearson depth correlation and dual-model consensus.
The total optimization objective $\mathcal{L}_{\text{total}}$ is defined as: 
\begin{equation}
\mathcal{L}_{\text{total}} 
= \mathcal{L}_{\text{train}}
+ \lambda_{\text{drift}}\,\mathcal{L}_{\text{anchor}}
+ \lambda_{\text{reg}}\,\mathcal{L}_{\text{reg}}
\end{equation}

\paragraph{Dynamic Training Loss}
This objective drives the primary learning of the 3D Gaussian field over the expanding trusted view space $\mathcal{V}_{\text{train}}$. Initially, $\mathcal{V}_{\text{train}}$ contains only the sparse observed input views. As training proceeds and reliable candidate viewpoints pass the verification criteria, they are progressively incorporated into $\mathcal{V}_{\text{train}}$. The dynamic reconstruction loss is computed over a sampled batch $\mathcal{B}_{\text{train}} \subset \mathcal{V}_{\text{train}}$:
\begin{equation}
\mathcal{L}_{\text{train}} = \frac{1}{|\mathcal{B}_{\text{train}}|} \sum_{I \in \mathcal{B}_{\text{train}}} \mathcal{L}_{\text{photo}}(I, \hat{I})
\end{equation}
where $\mathcal{L}_{\text{photo}}$ represents a combination of $\ell_1$ and D-SSIM losses. This dynamic supervision continuously broadens the angular constraints on Gaussian primitives, eliminating the under-constrained null-space responsible for ray-aligned overfitting.

\paragraph{Anchor Loss}
As the training pool $\mathcal{V}_{\text{train}}$ expands with anchored viewpoints, slight synthetic approximations could potentially accumulate over extended optimization, risking semantic or geometric drift. To ensure absolute fidelity to the true physical scene, we enforce a strict consistency constraint on the static, ground-truth observed views $\mathcal{V}_t$. At each optimization step, one original ground-truth view $I_{\text{anc}} \sim \mathcal{V}_t$ is randomly sampled and strictly revisited:
\begin{equation}
\mathcal{L}_{\text{anchor}} = \mathcal{L}_{\text{photo}}(I_{\text{anc}}, \hat{I}_{\text{anc}})
\end{equation}
By continuously revisiting real observations at every iteration, the anchor loss grounds the expanding 3D field to the physical camera coordinates, preventing accumulated drift without incurring the computational burden of rendering all observed views simultaneously.

\begin{table*}[!t]
\centering
\caption{Quantitative Comparison With State-of-the-Art Methods on the LLFF, MipNeRF-360, and DTU Datasets. FrontierGS Achieves the Best or Comparable Results Across PSNR, SSIM, and LPIPS Metrics Under Sparse-View Conditions.}
\label{tab:main_results}
\footnotesize 
\renewcommand{\arraystretch}{1.15}
\setlength{\tabcolsep}{8pt}
\resizebox{\textwidth}{!}{%
\begin{tabular}{l ccc ccc ccc}
\toprule
\multirow{2}{*}{Method} 
 & \multicolumn{3}{c}{LLFF (3 Views)} 
 & \multicolumn{3}{c}{MipNeRF-360 (24 Views)} 
 & \multicolumn{3}{c}{DTU (3 Views)} \\
\cmidrule(lr){2-4} \cmidrule(lr){5-7} \cmidrule(lr){8-10}
 & PSNR$\uparrow$ & SSIM$\uparrow$ & LPIPS$\downarrow$  
 & PSNR$\uparrow$ & SSIM$\uparrow$ & LPIPS$\downarrow$ 
 & PSNR$\uparrow$ & SSIM$\uparrow$ & LPIPS$\downarrow$ \\
\midrule
RegNeRF \cite{niemeyer2022regnerf}       & 19.08 & 0.587 & 0.336 & 22.19 & 0.643 & 0.335 & 18.89 & 0.745 & 0.190 \\
SparseNeRF \cite{wang2023sparsenerf}     & 19.86 & 0.624 & 0.328 & 22.85 & 0.693 & 0.315 & 19.55 & 0.769 & 0.201 \\
FreeNeRF \cite{yang2023freenerf}         & 19.63 & 0.612 & 0.308 & 22.78 & 0.689 & 0.323 & 19.92 & 0.787 & 0.182 \\
\midrule
3DGS \cite{kerbl20233d}                  & 18.54 & 0.588 & 0.272 & 21.71 & 0.672 & 0.248 & 17.65 & 0.816 & 0.146 \\
DNGaussian \cite{li2024dngaussian}       & 19.12 & 0.591 & 0.294 & 18.06 & 0.423 & 0.584 & 18.91 & 0.790 & 0.176 \\
FSGS \cite{zhu2024fsgs}                  & 20.43 & 0.682 & 0.248 & 23.40 & 0.733 & 0.238 & 17.14 & 0.818 & 0.162 \\
CoR-GS \cite{zhang2024cor}               & 20.45 & 0.712 & 0.196 & 23.55 & 0.727 & 0.226 & 19.21 & 0.853 & 0.119 \\
DropGaussian \cite{park2025dropgaussian} & 20.76 & 0.713 & 0.200 & 23.66 & 0.747 & 0.233 & 19.31 & 0.857 & 0.142 \\
NexusGS \cite{zheng2025nexusgs}          & 21.00 & 0.730 & \textbf{0.179} & 23.86 & 0.753 & 0.206 & 20.21 & 0.869 & \textbf{0.102} \\
LoopSparseGS \cite{bao2025loopsparsegs}  & 20.85 & 0.717 & 0.205 & 24.09 & 0.755 & 0.226 & \textbf{20.68} & 0.856 & 0.125 \\
\textbf{FrontierGS (Ours)}               & \textbf{21.10} & \textbf{0.732} & 0.193 & \textbf{24.21} & \textbf{0.761} & \textbf{0.202} & 20.45 & \textbf{0.873} & 0.129 \\
\bottomrule
\end{tabular}%
}
\end{table*}

\begin{figure*}[th]
    \centering    
    \includegraphics[width=1\linewidth]{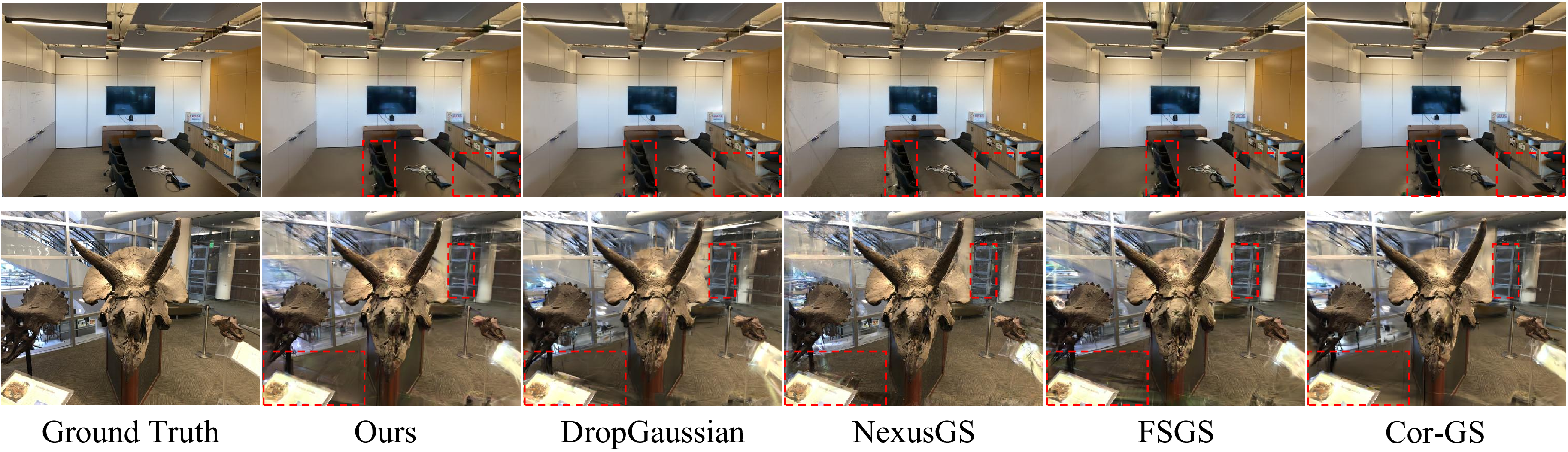}
    \caption{Qualitative comparison on the LLFF dataset under 3-view supervision. FrontierGS achieves sharper geometric contours, finer texture reconstruction, and reduced background tearing compared with baseline approaches.}
    \label{fig:llff}
\end{figure*}

\paragraph{Geometric Regularization Loss}
Because candidate viewpoints are virtual observations sampled along the expanding frontier, they initially lack ground-truth pixel supervision. Optimizing these viewpoints without geometric guidance could permit the formation of floating Gaussians and distorted surface contours. To resolve this, we impose two complementary regularization schemes directly on candidate views during training. 

The first scheme enforces geometry-aware structural alignment via Pearson depth correlation. Specifically, we employ a pretrained monocular depth estimation network \cite{ranftl2021vision} to predict a proxy depth map $D_{\text{proxy}}$ from the rendered candidate image $\hat{I}$. This proxy depth is compared against the metric depth map $D_{\text{render}}$ accumulated from 3DGS rasterization using the Pearson correlation coefficient:
\begin{equation}
    \mathcal{L}_{\text{depth}} = 1 - \frac{\text{Cov}(D_{\text{render}}, D_{\text{proxy}})}{\sigma_{\text{render}} \sigma_{\text{proxy}}}
\end{equation}
where $\text{Cov}(\cdot, \cdot)$ and $\sigma$ denote covariance and standard deviation, respectively. By measuring structural correlation rather than absolute depth values, $\mathcal{L}_{\text{depth}}$ guides the Gaussian primitives to form coherent spatial relief without suffering from the metric scale ambiguity or depth shift inherent to monocular estimators.

The second scheme introduces a dual-model consensus constraint. Valid scene geometry consistently converges across independently initialized models, whereas floating artifacts and degenerate Gaussian needles arise stochastically. Exploiting this consensus property, we maintain two independently initialized 3DGS models, $\mathcal{M}_A$ and $\mathcal{M}_B$. When rendering the candidate viewpoint, we penalize the photometric discrepancy between their respective outputs:
\begin{equation}
    \mathcal{L}_{\text{co}} = \| I_{\text{render}}^A - I_{\text{render}}^B \|_2^2
\end{equation}
This consensus loss drives both models toward structural agreement on true surfaces while suppressing uncoordinated stochastic floaters. The final geometric regularization term is formulated as the weighted sum:
\begin{equation}
\mathcal{L}_{\text{reg}} = \lambda_{\text{d}} \mathcal{L}_{\text{depth}} + \lambda_{\text{c}} \mathcal{L}_{\text{co}}
\end{equation}
where $\lambda_{\text{d}}$ and $\lambda_{\text{c}}$ balance the depth-correlation and consensus constraints, respectively.

\section{Experiments}
\label{sec:Experiments}

\subsection{Datasets}
To comprehensively evaluate the robustness and generalization ability of the proposed FrontierGS framework, we conduct extensive experiments across three diverse and widely adopted benchmarks: LLFF \cite{mildenhall2019local}, MipNeRF-360 \cite{barron2022mip}, and DTU \cite{jensen2014large}. These benchmarks encompass a broad range of challenging scenarios, including forward-facing complex scenes, unbounded $360^\circ$ captures with expansive depth ranges, and controlled object-centric captures. To ensure rigorous and fair comparisons, all dataset splits, image resolutions, and preprocessing protocols strictly follow established sparse-view evaluation standards \cite{zhu2024fsgs, zheng2025nexusgs, park2025dropgaussian, li2024dngaussian, zhang2024cor}.

\paragraph{LLFF} 
The LLFF benchmark \cite{mildenhall2019local} contains 8 forward-facing real-world scenes captured with handheld cameras. Following the standard protocol, every 8th image is held out as the novel test set, while the remaining views form the training candidate pool. Under the extreme sparse-view setting, we uniformly subsample 3 views across the trajectory as input supervision. All images are downsampled by a factor of 8 to balance computational efficiency and maintain consistency with baseline configurations.

\paragraph{MipNeRF-360} 
The MipNeRF-360 benchmark \cite{barron2022mip} consists of unbounded indoor and outdoor scenes characterized by complex 360-degree camera trajectories, wide spherical baselines, and dramatic depth ranges. This dataset thoroughly tests the ability of 3D representations to handle scale variations and background floater proliferation under sparse supervision. Consistent with prior sparse-view protocols, every 8th image is reserved for testing, and a sparse subset of 24 views is uniformly sampled from the remainder for training, with images downsampled by a factor of 8.

\paragraph{DTU} 
The DTU dataset \cite{jensen2014large} comprises object-centric laboratory scenes captured under controlled lighting conditions with precise camera pose annotations. Following common practice, our evaluation focuses on 15 representative scans (scans 8, 21, 30, 31, 34, 38, 40, 41, 45, 55, 63, 82, 103, 110, and 114). In the extreme 3-view setting, views 25, 22, and 28 are utilized for training, while the remaining 25 views serve as novel test viewpoints. Images are downsampled by a factor of 4, and binary object masks are employed to isolate foreground objects from background table artifacts.

\subsection{Implementation Details}
\label{subsec:implementation}

FrontierGS is implemented in PyTorch and built upon the official 3DGS codebase. All Gaussian primitive parameters are optimized using the Adam optimizer with identical learning rate schedules and densification policies as standard 3DGS \cite{kerbl20233d}. We set the anchor loss weight to $\lambda_{\text{drift}} = 1.0$, the Pearson depth correlation weight to $\lambda_{\text{d}} = 0.1$, and the dual-model consensus weight to $\lambda_{\text{c}} = 0.05$. For monocular depth guidance, a pretrained DPT model \cite{ranftl2021vision} is employed to predict proxy depth maps. The exploration radius naturally aligns with the intrinsic camera geometry of each capture scenario. For forward-facing scenes (LLFF) with narrow planar baselines, the exploration frontier spans up to an angular radius of $\sigma \le 10^\circ$ over 30,000 iterations. For unbounded $360^\circ$ environments (MipNeRF-360), exploration covers wider spherical baselines with fewer candidate probes per level to manage GPU memory. For object-centric captures (DTU), the exploration is naturally bounded within $\sigma \le 5^\circ$ over 13,000 iterations to ensure viewing frustums remain centered on the target subject, and standard foreground masks are applied following benchmark conventions. Complete schedules and reproducibility configurations are provided in the Supplementary Material.

\begin{figure*}[!t] 
    \centering    
    \includegraphics[width=1\linewidth]{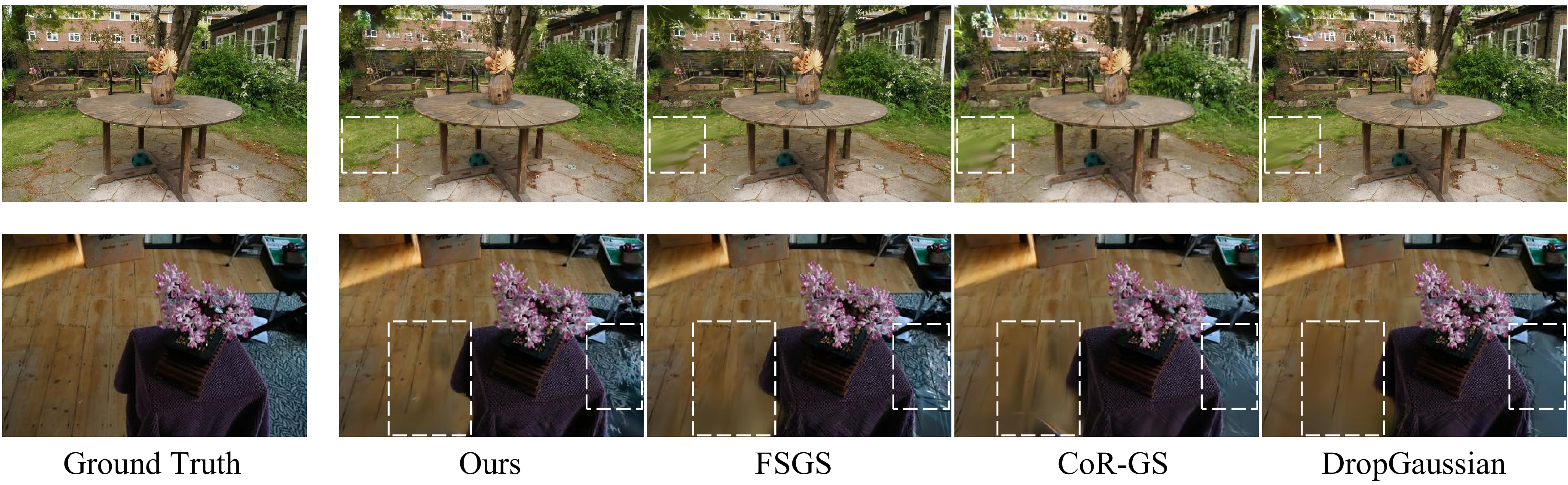}  
    \caption{Qualitative comparison on the MipNeRF-360 dataset. FrontierGS demonstrates superior perceptual fidelity and geometric stability across unbounded, large-scale scenes, suppressing distant cloudy floaters and unnatural color shifts.}
    \label{fig:mipnerf}
\end{figure*}
\begin{figure}[!t] 
    \centering    
    \includegraphics[width=1\linewidth]{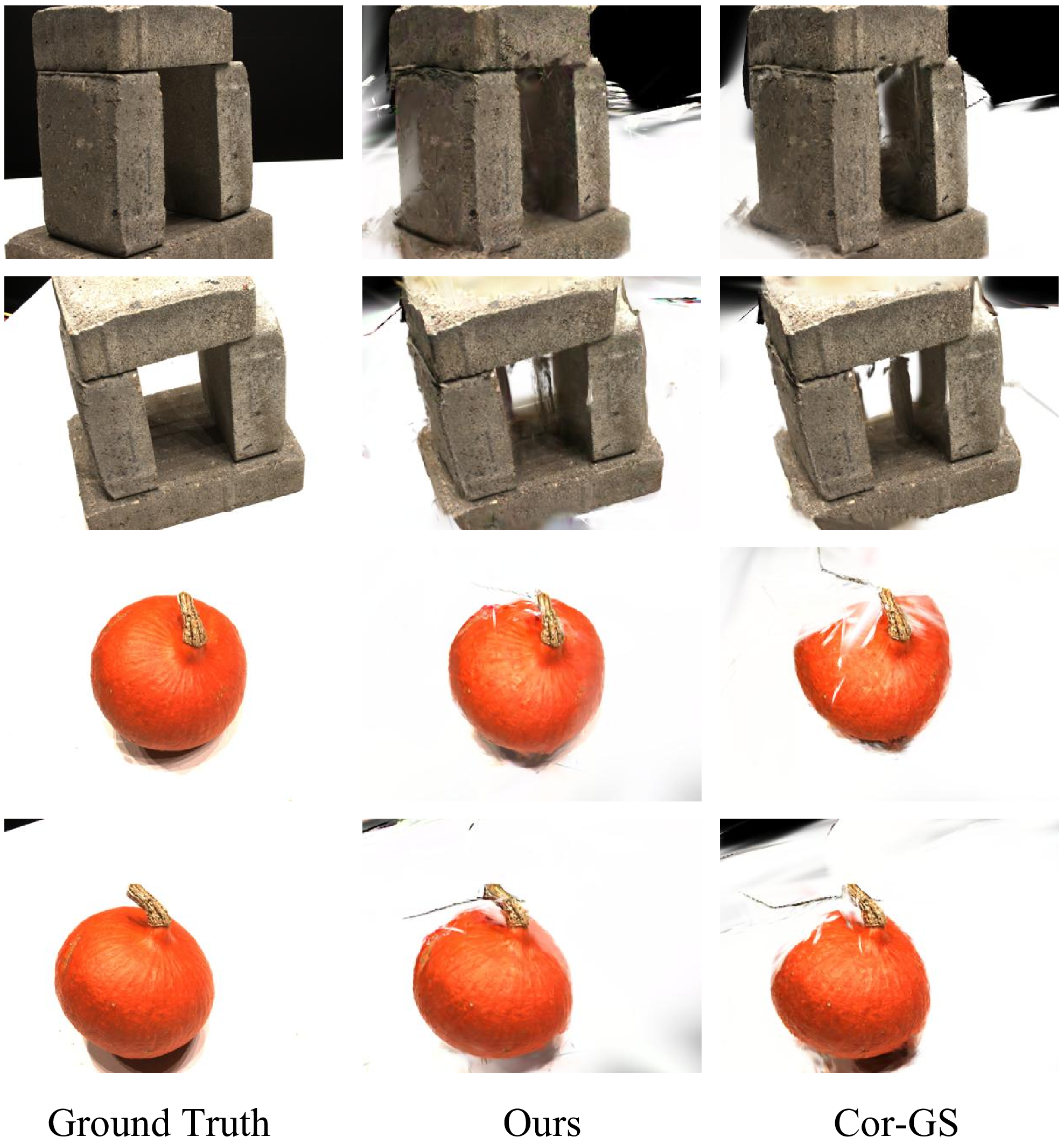}  
    \caption{Qualitative comparison on the DTU dataset under 3-view supervision. FrontierGS accurately preserves fine geometric contours and thin object boundaries compared to baseline methods.}
    \label{fig:dtu}
\end{figure}

\subsection{Baselines}
We compare FrontierGS against representative state-of-the-art sparse-view methods across two major methodological paradigms: NeRF-based regularizers including RegNeRF \cite{niemeyer2022regnerf}, SparseNeRF \cite{wang2023sparsenerf}, and FreeNeRF \cite{yang2023freenerf}; and 3DGS-based frameworks including Vanilla 3DGS \cite{kerbl20233d}, DNGaussian \cite{li2024dngaussian}, FSGS \cite{zhu2024fsgs}, CoR-GS \cite{zhang2024cor}, DropGaussian \cite{park2025dropgaussian}, NexusGS \cite{zheng2025nexusgs}, and LoopSparseGS \cite{bao2025loopsparsegs}. For quantitative performance assessment, we report standard novel-view synthesis metrics: Peak Signal-to-Noise Ratio (PSNR), SSIM, and LPIPS evaluated on held-out test viewpoints.

\subsection{Quantitative and Qualitative Results}

\paragraph{LLFF}
On forward-facing real-world scenes from the LLFF benchmark, FrontierGS achieves substantial improvements in both fidelity and perceptual realism, as documented in Tab.~\ref{tab:main_results}. Specifically, FrontierGS attains the highest PSNR (21.10 dB) and SSIM (0.732) among all compared 3DGS variants, outperforming vanilla 3DGS by +2.56 dB in PSNR and +0.144 in SSIM, while achieving LPIPS performance highly competitive with specialized baselines. As illustrated in the visual comparisons in Fig.~\ref{fig:llff}, these quantitative gains correspond to a marked reduction in cross-view texture smearing and floater artifacts. FrontierGS faithfully recovers sharp high-frequency details around intricate foliage boundaries and thin structures where baseline methods typically suffer from texture blurring or structural disintegration. This demonstrates that identifying and anchoring reliable intermediate states along the expanding frontier provides effective supervisory guidance that stabilizes 3D Gaussian geometry against sparse-view overfitting.

\paragraph{Mip-NeRF360}  
For large-scale unbounded scenes in MipNeRF-360, FrontierGS exhibits strong generalization and robust geometric coherence. Quantitatively, FrontierGS achieves the top performance among all compared methods across all evaluation metrics, yielding a PSNR of 24.21 dB, SSIM of 0.761, and LPIPS of 0.202 (Tab.~\ref{tab:main_results}). Qualitative comparisons are presented in Fig.~\ref{fig:mipnerf}. Compared to other baselines, FrontierGS substantially suppresses unnatural color shifts and hazy floating clusters in peripheral background areas, while accurately preserving fine textural details such as distant foliage and ground textures. These improvements confirm that progressively expanding the reliable supervision domain prevents peripheral Gaussians from dispersing into degenerate configurations, leading to coherent long-range geometry.

\begin{table}[!t]
\centering
\caption{Ablation study on the effect of progressive frontier expansion under varying view numbers across representative scenes of LLFF, MipNeRF-360, and DTU datasets.}
\label{tab:ablation_frontier}
\resizebox{\linewidth}{!}{%
\begin{tabular}{llccc}
\toprule
Dataset & \#Views / Setting & PSNR$\uparrow$ & SSIM$\uparrow$ & LPIPS$\downarrow$ \\
\midrule
\multirow{6}{*}{LLFF} 
 & 2 / Full & 20.94 & 0.768 & 0.194 \\
 & 2 / w/o Frontier Exp. & 18.30 & 0.601 & 0.231 \\
 & 3 / Full & 22.51 & 0.834 & 0.150 \\
 & 3 / w/o Frontier Exp. & 21.03 & 0.772 & 0.202 \\
 & 5 / Full & 23.95 & 0.861 & 0.140 \\
 & 5 / w/o Frontier Exp. & 22.97 & 0.841 & 0.143 \\
\midrule
\multirow{4}{*}{MipNeRF-360} 
 & 16 / Full & 21.23 & 0.781 & 0.189 \\
 & 16 / w/o Frontier Exp. & 20.32 & 0.754 & 0.202 \\
 & 24 / Full & 23.43 & 0.851 & 0.131 \\
 & 24 / w/o Frontier Exp. & 22.91 & 0.830 & 0.142 \\
\midrule
\multirow{6}{*}{DTU} 
 & 2 / Full & 18.66 & 0.922 & 0.077 \\
 & 2 / w/o Frontier Exp. & 15.04 & 0.896 & 0.092 \\
 & 3 / Full & 22.65 & 0.947 & 0.050 \\
 & 3 / w/o Frontier Exp. & 18.46 & 0.922 & 0.065 \\
 & 6 / Full & 24.45 & 0.947 & 0.043 \\
 & 6 / w/o Frontier Exp. & 20.51 & 0.946 & 0.048 \\
\bottomrule
\end{tabular}%
}
\end{table}

\paragraph{DTU}  
On the object-centric DTU benchmark under 3-view supervision, FrontierGS demonstrates distinct advantages in geometry-sensitive metrics, achieving the highest SSIM (0.873) along with competitive PSNR (20.45 dB) and LPIPS (0.129), as shown in Tab.~\ref{tab:main_results}. Qualitative comparisons in Fig.~\ref{fig:dtu} further highlight that FrontierGS preserves the overall structural integrity of the target objects more effectively, maintaining crisp surface boundaries, sharp geometric corners, and thin protrusions that are frequently smoothed out or partially eroded by other methods.

\begin{table}[!t]
\centering
\caption{Ablation study on the effect of loss components across representative scenes of LLFF, MipNeRF-360, and DTU datasets.}
\label{tab:ablation_loss}
\resizebox{\linewidth}{!}{%
\begin{tabular}{llccc}
\toprule
Dataset & \#Views / Setting & PSNR$\uparrow$ & SSIM$\uparrow$ & LPIPS$\downarrow$ \\
\midrule
\multirow{3}{*}{LLFF} 
 & 3 / Full & \textbf{22.51} & \textbf{0.834} & \textbf{0.150} \\
 & 3 / w/o Anchor Loss & 21.84 & 0.795 & 0.178 \\ 
 & 3 / w/o Candidate Reg. & 21.48 & 0.782 & 0.190 \\ 
\midrule
\multirow{3}{*}{MipNeRF-360} 
 & 16 / Full & \textbf{21.23} & \textbf{0.781} & \textbf{0.189} \\
 & 16 / w/o Anchor Loss & 20.88 & 0.768 & 0.194 \\ 
 & 16 / w/o Candidate Reg. & 20.65 & 0.760 & 0.198 \\ 
\midrule
\multirow{3}{*}{DTU} 
 & 3 / Full & \textbf{22.65} & \textbf{0.947} & \textbf{0.050} \\
 & 3 / w/o Anchor Loss & 21.97 & 0.935 & 0.058 \\ 
 & 3 / w/o Candidate Reg. & 20.82 & 0.929 & 0.061 \\ 
\bottomrule
\end{tabular}%
}
\end{table}

\begin{figure*}[!t] 
    \centering    
    \includegraphics[width=1\linewidth]{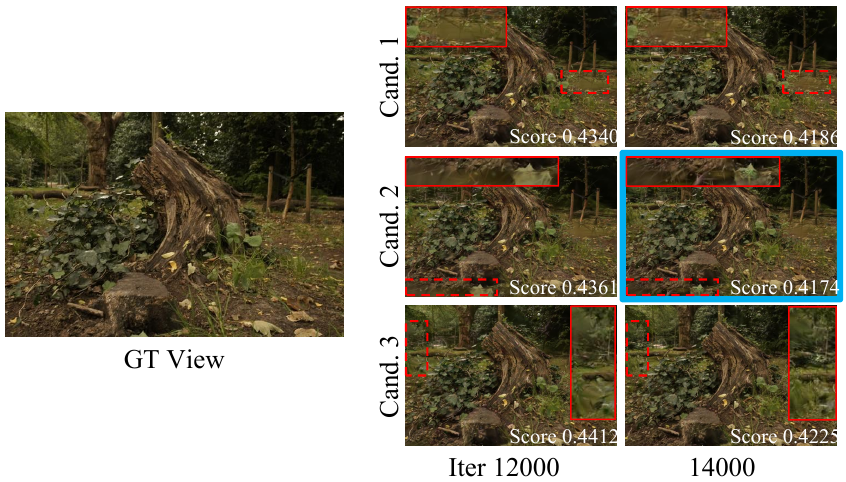} 
    \caption{Visualization of candidate viewpoint evolution and tracking under a specific perturbation exploration radius. Across progressive optimization stages, candidate renderings show systematic improvement in structural clarity and suppression of artifacts. The composite evaluation metric tracks candidate quality, with the optimal candidate passing the anchoring threshold highlighted by the blue bounding box.}
    \label{fig:stu}
\end{figure*}

\begin{figure*}[!t]
    \centering    
    \includegraphics[width=1\linewidth]{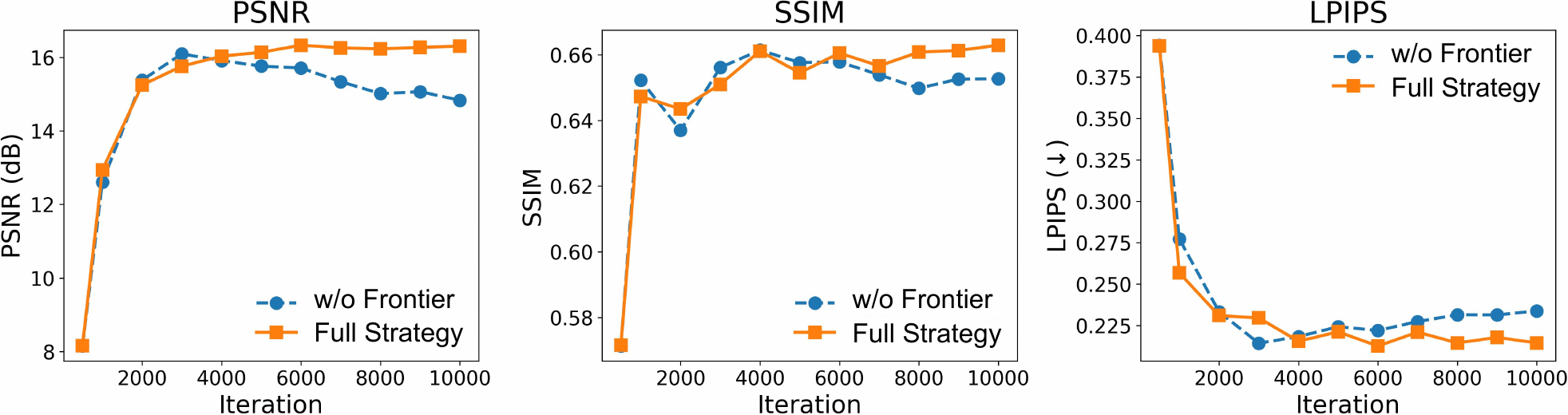}
    \caption{Evolution of PSNR, SSIM, and LPIPS during training on a representative scene of DTU with and without progressive frontier expansion.}
    \label{fig:aba}
\end{figure*}

\subsection{Ablation Study}
To thoroughly evaluate the individual contributions of our framework components, we perform extensive ablation studies across multiple view configurations and optimization loss terms on representative scenes of all benchmarks.

\paragraph{Effectiveness of Progressive Frontier Expansion}
The overall impact of progressive frontier expansion across varying datasets and input view numbers is validated in Tab.~\ref{tab:ablation_frontier}. Across all evaluated benchmarks, removing the frontier expansion mechanism causes consistent and substantial performance degradations across PSNR, SSIM, and LPIPS. On the LLFF dataset, under the extreme 2-view configuration, the full framework achieves a PSNR of 20.94 dB compared to 18.30 dB without frontier expansion, alongside a +0.167 increase in SSIM and a noticeable drop in LPIPS from 0.231 to 0.194. Similar significant performance advantages are maintained under 3-view and 5-view settings. On the MipNeRF-360 benchmark, incorporating frontier expansion improves PSNR from 20.32 dB to 21.23 dB at 16 views and from 22.91 dB to 23.43 dB at 24 views. On the DTU benchmark, the performance gains are even more pronounced, with the full strategy yielding a 3.5 to 4.2 dB improvement in PSNR over the baseline lacking frontier expansion (e.g., 22.65 dB vs. 18.46 dB at 3 views). These consistent gains verify that progressive expansion provides indispensable supervision across diverse scene types and view densities.

\paragraph{Effectiveness of Loss Components}
We examine the structural roles of our core optimization loss terms in Tab.~\ref{tab:ablation_loss}: the dynamic reconstruction loss, the anchor loss on ground-truth observations, and the candidate geometric regularizers. Removing the anchor loss $\mathcal{L}_{\text{anchor}}$ results in noticeable performance drops across all datasets (e.g., $-0.67$ dB on LLFF and $-0.68$ dB on DTU), confirming its crucial role in grounding the expanding 3D field to true physical observations and preventing accumulated coordinate drift. Excluding candidate geometric regularizations $\mathcal{L}_{\text{reg}}$ leads to severe degradation ($-1.03$ dB on LLFF and $-1.83$ dB on DTU), as unconstrained candidate viewpoints introduce floating density and irregular geometry that corrupt optimization. The full formulation ensures that dynamic supervision expands safely while maintaining strict fidelity to observed geometry.

\subsection{Analysis of Viewpoint Verification and Anchoring Mechanism}
In sparse-view reconstruction, the central bottleneck stems from severe supervision scarcity across the camera manifold. While introducing virtual viewpoints helps bridge these supervision voids, indiscriminately admitting sub-optimally rendered views inevitably injects corrupt geometry and reinforces training artifacts. Therefore, establishing a rigorous, parallax-aware criterion to evaluate and anchor candidate viewpoints is critical.

To provide reliable quality assessment in the absence of ground-truth novel-view imagery, we employ a multi-signal evaluation metric during the continuous tracking phase. This metric combines full-reference metrics (SSIM and LPIPS) with a no-reference image naturalness score. SSIM and LPIPS assess structural and perceptual fidelity relative to adjacent observed cameras, maintaining spatial coherence with verified physical observations. Concurrently, the no-reference score penalizes unnatural rendering noise, high-frequency blurring, and floating debris, providing an effective proxy for standalone visual plausibility.

Crucially, when transitioning between perturbation radius stages, the criteria adaptively decouple by relying solely on the no-reference quality score for permanent anchoring. Enforcing reference-based SSIM and LPIPS at this juncture would unfairly penalize candidate viewpoints with wider baselines due to genuine perspective parallax, inherently biasing selection toward viewpoints clustered tightly around the original cameras and restricting viewpoint diversity. By accepting valid geometric parallax while strictly enforcing high standalone visual naturalness, the framework successfully admits diverse, structurally sound viewpoints. As visualized in Fig.~\ref{fig:stu}, this dynamic mechanism steadily refines candidate renders and selects the optimal representation that balances spatial diversity with geometric coherence.

\subsection{Analysis of Training Dynamics}
To empirically investigate how progressive frontier expansion resolves sparse-view overfitting, we monitor the trajectory of PSNR, SSIM, and LPIPS metrics throughout optimization on a representative scene of DTU. As depicted in Fig.~\ref{fig:aba}, the baseline optimization without frontier expansion exhibits a clear performance decay during intermediate-to-late training—a direct manifestation of view-space generalization collapse, where the model memorizes observed rays at the cost of global 3D geometry. In stark contrast, FrontierGS maintains a consistently ascending performance trajectory across all metrics throughout training. By actively identifying, regularizing, and permanently anchoring transiently reliable representations along the expanding frontier, FrontierGS converts fragile intermediate generalization into enduring supervision, ensuring stable convergence toward high-fidelity 3D reconstruction.

\section{Conclusion}
In this paper, we presented FrontierGS, a progressive view-space expansion framework for sparse-view 3D Gaussian Splatting. Addressing the fundamental pathology of view-space generalization collapse—where novel-view quality peaks early before over-optimization along observed rays destroys global 3D coherence—FrontierGS recasts sparse reconstruction into an active paradigm of probing, verifying, and anchoring transiently reliable states along a dynamic generalization frontier. Probed candidate viewpoints are regularized by Pearson depth correlation and dual-model consensus during training, while a parallax-aware multi-signal metric screens and permanently anchors high-fidelity states into the supervision pool before degradation occurs. Extensive experiments across the LLFF, MipNeRF-360, and DTU benchmarks demonstrate that FrontierGS consistently outperforms state-of-the-art NeRF- and 3DGS-based methods in rendering fidelity and geometric consistency while preserving native real-time rendering speeds.

\bibliographystyle{elsarticle-harv}
\bibliography{references}

%% file: main_arxiv.bbl
\begin{thebibliography}{51}
\providecommand{\natexlab}[1]{#1}
\providecommand{\url}[1]{\texttt{#1}}
\expandafter\ifx\csname urlstyle\endcsname\relax
  \providecommand{\doi}[1]{doi: #1}\else
  \providecommand{\doi}{doi: \begingroup \urlstyle{rm}\Url}\fi

\bibitem[Agnolucci et~al.(2024)Agnolucci, Galteri, and
  Bertini]{agnolucci2024qualityaware}
Lorenzo Agnolucci, Leonardo Galteri, and Marco Bertini.
\newblock Quality-aware image-text alignment for opinion-unaware image quality
  assessment.
\newblock \emph{arXiv preprint arXiv:2403.11176}, 2024.

\bibitem[Bao et~al.(2025)Bao, Liao, Zhou, Liu, Li, and
  Qiu]{bao2025loopsparsegs}
Zhenyu Bao, Guibiao Liao, Kaichen Zhou, Kanglin Liu, Qing Li, and Guoping Qiu.
\newblock Loopsparsegs: Loop based sparse-view friendly gaussian splatting.
\newblock \emph{IEEE Transactions on Image Processing}, 2025.

\bibitem[Barron et~al.(2021)Barron, Mildenhall, Tancik, Hedman, Martin-Brualla,
  and Srinivasan]{barron2021mip}
Jonathan~T Barron, Ben Mildenhall, Matthew Tancik, Peter Hedman, Ricardo
  Martin-Brualla, and Pratul~P Srinivasan.
\newblock Mip-nerf: A multiscale representation for anti-aliasing neural
  radiance fields.
\newblock In \emph{Proceedings of the IEEE/CVF international conference on
  computer vision}, pages 5855--5864, 2021.

\bibitem[Barron et~al.(2022)Barron, Mildenhall, Verbin, Srinivasan, and
  Hedman]{barron2022mip}
Jonathan~T Barron, Ben Mildenhall, Dor Verbin, Pratul~P Srinivasan, and Peter
  Hedman.
\newblock Mip-nerf 360: Unbounded anti-aliased neural radiance fields.
\newblock In \emph{Proceedings of the IEEE/CVF conference on computer vision
  and pattern recognition}, pages 5470--5479, 2022.

\bibitem[Cai et~al.(2024)Cai, Wang, Yuille, Zhou, and Wang]{cai2024structure}
Yuanhao Cai, Jiahao Wang, Alan Yuille, Zongwei Zhou, and Angtian Wang.
\newblock Structure-aware sparse-view x-ray 3d reconstruction.
\newblock In \emph{Proceedings of the IEEE/CVF conference on computer vision
  and pattern recognition}, pages 11174--11183, 2024.

\bibitem[Charatan et~al.(2024)Charatan, Li, Tagliasacchi, and
  Sitzmann]{charatan2024pixelsplat}
David Charatan, Sizhe~Lester Li, Andrea Tagliasacchi, and Vincent Sitzmann.
\newblock pixelsplat: 3d gaussian splats from image pairs for scalable
  generalizable 3d reconstruction.
\newblock In \emph{Proceedings of the IEEE/CVF conference on computer vision
  and pattern recognition}, pages 19457--19467, 2024.

\bibitem[Chen et~al.(2021)Chen, Xu, Zhao, Zhang, Xiang, Yu, and
  Su]{chen2021mvsnerf}
Anpei Chen, Zexiang Xu, Fuqiang Zhao, Xiaoshuai Zhang, Fanbo Xiang, Jingyi Yu,
  and Hao Su.
\newblock Mvsnerf: Fast generalizable radiance field reconstruction from
  multi-view stereo.
\newblock In \emph{Proceedings of the IEEE/CVF international conference on
  computer vision}, pages 14124--14133, 2021.

\bibitem[Dai et~al.(2025)Dai, Tang, Zhang, and Xie]{mgss}
Zhien Dai, Zhaohui Tang, Hu~Zhang, and Yongfang Xie.
\newblock Mgs-stereo: Multi-scale geometric-structure-enhanced stereo matching
  for complex real-world scenes.
\newblock \emph{IEEE Transactions on Image Processing}, 34:\penalty0
  6246--6258, 2025.

\bibitem[Fei et~al.(2026)Fei, Gao, Hu, Hou, Li, Ren, and Zhang]{FEI2026130758}
Sicheng Fei, Xuehao Gao, Jinwen Hu, Xiaolei Hou, Lei Li, Jun Ren, and Dingwen
  Zhang.
\newblock Knowledge distillation-based distributed dynamic 3d gaussian
  splatting for large scale scene reconstruction.
\newblock \emph{Expert Systems with Applications}, 305:\penalty0 130758, 2026.
\newblock ISSN 0957-4174.

\bibitem[Han et~al.(2024)Han, Zhou, Liu, and Han]{han2024binocular}
Liang Han, Junsheng Zhou, Yu-Shen Liu, and Zhizhong Han.
\newblock Binocular-guided 3d gaussian splatting with view consistency for
  sparse view synthesis.
\newblock \emph{Advances in Neural Information Processing Systems},
  37:\penalty0 68595--68621, 2024.

\bibitem[Jensen et~al.(2014)Jensen, Dahl, Vogiatzis, Tola, and
  Aan{\ae}s]{jensen2014large}
Rasmus Jensen, Anders Dahl, George Vogiatzis, Engin Tola, and Henrik Aan{\ae}s.
\newblock Large scale multi-view stereopsis evaluation.
\newblock In \emph{Proceedings of the IEEE conference on computer vision and
  pattern recognition}, pages 406--413, 2014.

\bibitem[Kerbl et~al.(2023)Kerbl, Kopanas, Leimk{\"u}hler, and
  Drettakis]{kerbl20233d}
Bernhard Kerbl, Georgios Kopanas, Thomas Leimk{\"u}hler, and George Drettakis.
\newblock 3d gaussian splatting for real-time radiance field rendering.
\newblock \emph{ACM Trans. Graph.}, 42\penalty0 (4):\penalty0 139--1, 2023.

\bibitem[Kong et~al.(2025)Kong, Yang, and Wang]{11095131}
Hanyang Kong, Xingyi Yang, and Xinchao Wang.
\newblock Generative sparse-view gaussian splatting.
\newblock In \emph{2025 IEEE/CVF Conference on Computer Vision and Pattern
  Recognition (CVPR)}, pages 26745--26755, 2025.

\bibitem[Li et~al.(2024)Li, Zhang, Bai, Zheng, Ning, Zhou, and
  Gu]{li2024dngaussian}
Jiahe Li, Jiawei Zhang, Xiao Bai, Jin Zheng, Xin Ning, Jun Zhou, and Lin Gu.
\newblock Dngaussian: Optimizing sparse-view 3d gaussian radiance fields with
  global-local depth normalization.
\newblock In \emph{Proceedings of the IEEE/CVF conference on computer vision
  and pattern recognition}, pages 20775--20785, 2024.

\bibitem[Liang et~al.(2026)Liang, Liu, Li, Shen, and Yao]{LIANG2026133170}
Jiulin Liang, Caixia Liu, Haisheng Li, Wenshan Shen, and Qianwen Yao.
\newblock 3d gaussian splatting for reconstruction: Methods, datasets, and
  comparisons.
\newblock \emph{Expert Systems with Applications}, 331:\penalty0 133170, 2026.
\newblock ISSN 0957-4174.

\bibitem[Liu et~al.(2026{\natexlab{a}})Liu, Liang, Chen, Wang, and
  Na]{LIU2026129334}
Ming Liu, Yuxuan Liang, Siwei Chen, Junjie Wang, and Yang Na.
\newblock Sv-2dgs: Optimization of sparse view 3d reconstruction based on 2dgs
  models.
\newblock \emph{Expert Systems with Applications}, 297:\penalty0 129334,
  2026{\natexlab{a}}.
\newblock ISSN 0957-4174.

\bibitem[Liu et~al.(2026{\natexlab{b}})Liu, Liang, Chen, Wang, and
  Na]{liu2026sv}
Ming Liu, Yuxuan Liang, Siwei Chen, Junjie Wang, and Yang Na.
\newblock Sv-2dgs: Optimization of sparse view 3d reconstruction based on 2dgs
  models.
\newblock \emph{Expert Systems with Applications}, 297:\penalty0 129334,
  2026{\natexlab{b}}.

\bibitem[Liu et~al.(2024)Liu, Luo, Fan, Wang, Peng, and
  Zhang]{liu2024citygaussian}
Yang Liu, Chuanchen Luo, Lue Fan, Naiyan Wang, Junran Peng, and Zhaoxiang
  Zhang.
\newblock Citygaussian: Real-time high-quality large-scale scene rendering with
  gaussians.
\newblock In \emph{European Conference on Computer Vision}, pages 265--282.
  Springer, 2024.

\bibitem[Long et~al.(2022)Long, Lin, Wang, Komura, and
  Wang]{long2022sparseneus}
Xiaoxiao Long, Cheng Lin, Peng Wang, Taku Komura, and Wenping Wang.
\newblock Sparseneus: Fast generalizable neural surface reconstruction from
  sparse views.
\newblock In \emph{European Conference on Computer Vision}, pages 210--227.
  Springer, 2022.

\bibitem[Mallick et~al.(2024)Mallick, Goel, Kerbl, Steinberger, Carrasco, and
  De~La~Torre]{mallick2024taming}
Saswat~Subhajyoti Mallick, Rahul Goel, Bernhard Kerbl, Markus Steinberger,
  Francisco~Vicente Carrasco, and Fernando De~La~Torre.
\newblock Taming 3dgs: High-quality radiance fields with limited resources.
\newblock In \emph{SIGGRAPH Asia 2024 Conference Papers}, pages 1--11, 2024.

\bibitem[Mildenhall et~al.(2019)Mildenhall, Srinivasan, Ortiz-Cayon, Kalantari,
  Ramamoorthi, Ng, and Kar]{mildenhall2019local}
Ben Mildenhall, Pratul~P Srinivasan, Rodrigo Ortiz-Cayon, Nima~Khademi
  Kalantari, Ravi Ramamoorthi, Ren Ng, and Abhishek Kar.
\newblock Local light field fusion: Practical view synthesis with prescriptive
  sampling guidelines.
\newblock \emph{ACM Transactions on Graphics (ToG)}, 38\penalty0 (4):\penalty0
  1--14, 2019.

\bibitem[Mildenhall et~al.(2021)Mildenhall, Srinivasan, Tancik, Barron,
  Ramamoorthi, and Ng]{mildenhall2021nerf}
Ben Mildenhall, Pratul~P Srinivasan, Matthew Tancik, Jonathan~T Barron, Ravi
  Ramamoorthi, and Ren Ng.
\newblock Nerf: Representing scenes as neural radiance fields for view
  synthesis.
\newblock \emph{Communications of the ACM}, 65\penalty0 (1):\penalty0 99--106,
  2021.

\bibitem[Niemeyer et~al.(2022)Niemeyer, Barron, Mildenhall, Sajjadi, Geiger,
  and Radwan]{niemeyer2022regnerf}
Michael Niemeyer, Jonathan~T Barron, Ben Mildenhall, Mehdi~SM Sajjadi, Andreas
  Geiger, and Noha Radwan.
\newblock Regnerf: Regularizing neural radiance fields for view synthesis from
  sparse inputs.
\newblock In \emph{Proceedings of the IEEE/CVF conference on computer vision
  and pattern recognition}, pages 5480--5490, 2022.

\bibitem[Paliwal et~al.(2024)Paliwal, Ye, Xiong, Kotovenko, Ranjan, Chandra,
  and Kalantari]{paliwal2024coherentgs}
Avinash Paliwal, Wei Ye, Jinhui Xiong, Dmytro Kotovenko, Rakesh Ranjan, Vikas
  Chandra, and Nima~Khademi Kalantari.
\newblock Coherentgs: Sparse novel view synthesis with coherent 3d gaussians.
\newblock In \emph{European Conference on Computer Vision}, pages 19--37.
  Springer, 2024.

\bibitem[Park et~al.(2025)Park, Ryu, and Kim]{park2025dropgaussian}
Hyunwoo Park, Gun Ryu, and Wonjun Kim.
\newblock Dropgaussian: Structural regularization for sparse-view gaussian
  splatting.
\newblock In \emph{Proceedings of the Computer Vision and Pattern Recognition
  Conference}, pages 21600--21609, 2025.

\bibitem[Pumarola et~al.(2021)Pumarola, Corona, Pons-Moll, and
  Moreno-Noguer]{pumarola2021d}
Albert Pumarola, Enric Corona, Gerard Pons-Moll, and Francesc Moreno-Noguer.
\newblock D-nerf: Neural radiance fields for dynamic scenes.
\newblock In \emph{Proceedings of the IEEE/CVF conference on computer vision
  and pattern recognition}, pages 10318--10327, 2021.

\bibitem[Qi et~al.(2017)Qi, Su, Mo, and Guibas]{qi2017pointnet}
Charles~R Qi, Hao Su, Kaichun Mo, and Leonidas~J Guibas.
\newblock Pointnet: Deep learning on point sets for 3d classification and
  segmentation.
\newblock In \emph{Proceedings of the IEEE conference on computer vision and
  pattern recognition}, pages 652--660, 2017.

\bibitem[Qian et~al.(2024)Qian, Wang, Mihajlovic, Geiger, and
  Tang]{qian20243dgs}
Zhiyin Qian, Shaofei Wang, Marko Mihajlovic, Andreas Geiger, and Siyu Tang.
\newblock 3dgs-avatar: Animatable avatars via deformable 3d gaussian splatting.
\newblock In \emph{Proceedings of the IEEE/CVF conference on computer vision
  and pattern recognition}, pages 5020--5030, 2024.

\bibitem[Ranftl et~al.(2021)Ranftl, Bochkovskiy, and Koltun]{ranftl2021vision}
Ren{\'e} Ranftl, Alexey Bochkovskiy, and Vladlen Koltun.
\newblock Vision transformers for dense prediction.
\newblock In \emph{Proceedings of the IEEE/CVF international conference on
  computer vision}, pages 12179--12188, 2021.

\bibitem[Schonberger and Frahm(2016)]{schonberger2016structure}
Johannes~L Schonberger and Jan-Michael Frahm.
\newblock Structure-from-motion revisited.
\newblock In \emph{Proceedings of the IEEE conference on computer vision and
  pattern recognition}, pages 4104--4113, 2016.

\bibitem[Skublewska-Paszkowska et~al.(2022)Skublewska-Paszkowska, Milosz,
  Powroznik, and Lukasik]{skublewska20223d}
Maria Skublewska-Paszkowska, Marek Milosz, Pawel Powroznik, and Edyta Lukasik.
\newblock 3d technologies for intangible cultural heritage
  preservation—literature review for selected databases.
\newblock \emph{Heritage Science}, 10\penalty0 (1):\penalty0 3, 2022.

\bibitem[Somraj and Soundararajan(2023)]{somraj2023vip}
Nagabhushan Somraj and Rajiv Soundararajan.
\newblock Vip-nerf: Visibility prior for sparse input neural radiance fields.
\newblock In \emph{ACM SIGGRAPH 2023 conference proceedings}, pages 1--11,
  2023.

\bibitem[Szymanowicz et~al.(2024)Szymanowicz, Rupprecht, and
  Vedaldi]{szymanowicz2024splatter}
Stanislaw Szymanowicz, Chrisitian Rupprecht, and Andrea Vedaldi.
\newblock Splatter image: Ultra-fast single-view 3d reconstruction.
\newblock In \emph{Proceedings of the IEEE/CVF conference on computer vision
  and pattern recognition}, pages 10208--10217, 2024.

\bibitem[Tang et~al.(2025)Tang, Guo, Li, and Peng]{tang2025spars3r}
Yutao Tang, Yuxiang Guo, Deming Li, and Cheng Peng.
\newblock Spars3r: Semantic prior alignment and regularization for sparse 3d
  reconstruction.
\newblock In \emph{Proceedings of the Computer Vision and Pattern Recognition
  Conference}, pages 26810--26821, 2025.

\bibitem[Tao et~al.(2022)Tao, Xiao, Qi, Cheng, and Ji]{tao2022digital}
Fei Tao, Bin Xiao, Qinglin Qi, Jiangfeng Cheng, and Ping Ji.
\newblock Digital twin modeling.
\newblock \emph{Journal of Manufacturing Systems}, 64:\penalty0 372--389, 2022.

\bibitem[Wang et~al.(2023)Wang, Chen, Loy, and Liu]{wang2023sparsenerf}
Guangcong Wang, Zhaoxi Chen, Chen~Change Loy, and Ziwei Liu.
\newblock Sparsenerf: Distilling depth ranking for few-shot novel view
  synthesis.
\newblock In \emph{Proceedings of the IEEE/CVF international conference on
  computer vision}, pages 9065--9076, 2023.

\bibitem[Wang et~al.(2025{\natexlab{a}})Wang, Wang, Yang, and Zhang]{clocap}
Kangkan Wang, Chong Wang, Jian Yang, and Guofeng Zhang.
\newblock Clocap-gs: Clothed human performance capture with 3d gaussian
  splatting.
\newblock \emph{IEEE Transactions on Image Processing}, 34:\penalty0
  5200--5214, 2025{\natexlab{a}}.
\newblock \doi{10.1109/TIP.2025.3592534}.

\bibitem[Wang et~al.(2018)Wang, Zhang, Li, Fu, Liu, and
  Jiang]{wang2018pixel2mesh}
Nanyang Wang, Yinda Zhang, Zhuwen Li, Yanwei Fu, Wei Liu, and Yu-Gang Jiang.
\newblock Pixel2mesh: Generating 3d mesh models from single rgb images.
\newblock In \emph{Proceedings of the European conference on computer vision
  (ECCV)}, pages 52--67, 2018.

\bibitem[Wang et~al.(2025{\natexlab{b}})Wang, Wei, Lu, and Kang]{plgs}
Yu~Wang, Xiaobao Wei, Ming Lu, and Guoliang Kang.
\newblock Plgs: Robust panoptic lifting with 3d gaussian splatting.
\newblock \emph{IEEE Transactions on Image Processing}, 34:\penalty0
  3377--3388, 2025{\natexlab{b}}.

\bibitem[Wang et~al.(2024)Wang, Yi, Wu, Zhao, Chen, and Zhang]{wang2024view}
Yuxuan Wang, Xuanyu Yi, Zike Wu, Na~Zhao, Long Chen, and Hanwang Zhang.
\newblock View-consistent 3d editing with gaussian splatting.
\newblock In \emph{European conference on computer vision}, pages 404--420.
  Springer, 2024.

\bibitem[Wu et~al.(2024)Wu, Mildenhall, Henzler, Park, Gao, Watson, Srinivasan,
  Verbin, Barron, Poole, and Hołyński]{10657150}
Rundi Wu, Ben Mildenhall, Philipp Henzler, Keunhong Park, Ruiqi Gao, Daniel
  Watson, Pratul~P. Srinivasan, Dor Verbin, Jonathan~T. Barron, Ben Poole, and
  Aleksander Hołyński.
\newblock Reconfusion: 3d reconstruction with diffusion priors.
\newblock In \emph{2024 IEEE/CVF Conference on Computer Vision and Pattern
  Recognition (CVPR)}, pages 21551--21561, 2024.

\bibitem[Xiong et~al.(2025)Xiong, Muttukuru, Xiao, Upadhyay, Chari, Zhao, and
  Kadambi]{xiong2025sparsegs}
Haolin Xiong, Sairisheek Muttukuru, Hanyuan Xiao, Rishi Upadhyay, Pradyumna
  Chari, Yajie Zhao, and Achuta Kadambi.
\newblock Sparsegs: Sparse view synthesis using 3d gaussian splatting.
\newblock In \emph{2025 International Conference on 3D Vision (3DV)}, pages
  1032--1041. IEEE, 2025.

\bibitem[Xu et~al.(2024)Xu, Shi, Yifan, Chen, Yang, Peng, Shen, and
  Wetzstein]{xu2024grm}
Yinghao Xu, Zifan Shi, Wang Yifan, Hansheng Chen, Ceyuan Yang, Sida Peng, Yujun
  Shen, and Gordon Wetzstein.
\newblock Grm: Large gaussian reconstruction model for efficient 3d
  reconstruction and generation.
\newblock In \emph{European Conference on Computer Vision}, pages 1--20.
  Springer, 2024.

\bibitem[Yang et~al.(2023)Yang, Pavone, and Wang]{yang2023freenerf}
Jiawei Yang, Marco Pavone, and Yue Wang.
\newblock Freenerf: Improving few-shot neural rendering with free frequency
  regularization.
\newblock In \emph{Proceedings of the IEEE/CVF conference on computer vision
  and pattern recognition}, pages 8254--8263, 2023.

\bibitem[Yu et~al.(2021)Yu, Ye, Tancik, and Kanazawa]{yu2021pixelnerf}
Alex Yu, Vickie Ye, Matthew Tancik, and Angjoo Kanazawa.
\newblock pixelnerf: Neural radiance fields from one or few images.
\newblock In \emph{Proceedings of the IEEE/CVF conference on computer vision
  and pattern recognition}, pages 4578--4587, 2021.

\bibitem[Yu et~al.(2024)Yu, Liu, Tan, Wang, and Zhang]{yu2024sensors}
Shuwan Yu, Xiaoang Liu, Qianqiu Tan, Zitong Wang, and Baohua Zhang.
\newblock Sensors, systems and algorithms of 3d reconstruction for smart
  agriculture and precision farming: A review.
\newblock \emph{Computers and Electronics in Agriculture}, 224:\penalty0
  109229, 2024.

\bibitem[Zhang et~al.(2025)Zhang, Gajardo, Medic, Katircioglu, Boss,
  Kirchgessner, Walter, and Roth]{zhang2025wheat3dgs}
Daiwei Zhang, Joaquin Gajardo, Tomislav Medic, Isinsu Katircioglu, Mike Boss,
  Norbert Kirchgessner, Achim Walter, and Lukas Roth.
\newblock Wheat3dgs: In-field 3d reconstruction, instance segmentation and
  phenotyping of wheat heads with gaussian splatting.
\newblock In \emph{Proceedings of the Computer Vision and Pattern Recognition
  Conference}, pages 5360--5370, 2025.

\bibitem[Zhang et~al.(2024)Zhang, Li, Yu, Huang, Gu, Zheng, and
  Bai]{zhang2024cor}
Jiawei Zhang, Jiahe Li, Xiaohan Yu, Lei Huang, Lin Gu, Jin Zheng, and Xiao Bai.
\newblock Cor-gs: sparse-view 3d gaussian splatting via co-regularization.
\newblock In \emph{European Conference on Computer Vision}, pages 335--352.
  Springer, 2024.

\bibitem[Zhang et~al.(2026)Zhang, Wei, Wang, and Liu]{zhang2026usgs}
Yao Zhang, Jiangshu Wei, Yuchao Wang, and Jiajun Liu.
\newblock Usgs: Enhancing sparse view synthesis with unseen viewpoint
  regularization in 3d gaussian splatting.
\newblock \emph{Pattern Recognition}, 170:\penalty0 112087, 2026.

\bibitem[Zheng et~al.(2025)Zheng, Jiang, He, Sun, Dong, Zhang, and
  Du]{zheng2025nexusgs}
Yulong Zheng, Zicheng Jiang, Shengfeng He, Yandu Sun, Junyu Dong, Huaidong
  Zhang, and Yong Du.
\newblock Nexusgs: Sparse view synthesis with epipolar depth priors in 3d
  gaussian splatting.
\newblock In \emph{Proceedings of the Computer Vision and Pattern Recognition
  Conference}, pages 26800--26809, 2025.

\bibitem[Zhu et~al.(2024)Zhu, Fan, Jiang, and Wang]{zhu2024fsgs}
Zehao Zhu, Zhiwen Fan, Yifan Jiang, and Zhangyang Wang.
\newblock Fsgs: Real-time few-shot view synthesis using gaussian splatting.
\newblock In \emph{European conference on computer vision}, pages 145--163.
  Springer, 2024.

\end{thebibliography}
